\pdfoutput=1

\documentclass[11pt]{article}

\usepackage[preprint]{acl}

\usepackage{times}
\usepackage{latexsym}
\usepackage{makecell}
\usepackage{longtable} %
\usepackage[T1]{fontenc}

\usepackage[utf8]{inputenc}

\usepackage{microtype}

\usepackage{inconsolata}

\usepackage{graphicx}

\usepackage{todonotes}
\usepackage{comment}
\usepackage{enumitem}

\title{Reasoning and the Trusting Behavior of DeepSeek and GPT: \linebreak An Experiment Revealing Hidden Fault Lines in Large Language Models}

\author{Rubing Li \\
  New York University \\
  Stern School of Business \\
  \texttt{rl4229@stern.nyu.edu} \\\And
  Jo\~{a}o Sedoc \\
  New York University \\
  Stern School of Business \\
  \texttt{jsedoc@stern.nyu.edu} \\\And
  Arun Sundararajan \\
  New York University \\
  Stern School of Business \\
  \texttt{digitalarun@nyu.edu} \\}

\begin{document}
\maketitle
\begin{abstract}
When encountering increasingly frequent performance improvements or cost reductions from a new large language model (LLM), developers of applications leveraging LLMs must decide whether to take advantage of these improvements or stay with older tried-and-tested models. Low perceived switching frictions can lead to choices that do not consider more subtle behavior changes that the transition may induce. Our experiments use a popular game-theoretic behavioral economics model of trust to show stark differences in the trusting behavior of OpenAI's and DeepSeek's models. We highlight a collapse in the economic trust behavior of the o1-mini and o3-mini models as they reconcile profit-maximizing and risk-seeking with future returns from trust, and contrast it with DeepSeek's more sophisticated and profitable trusting behavior that stems from an ability to incorporate deeper concepts like forward planning and theory-of-mind. As LLMs form the basis for high-stakes commercial systems, our results highlight the perils of relying on LLM performance benchmarks that are too narrowly defined and suggest that careful analysis of their hidden fault lines should be part of any organization's AI strategy.   
\end{abstract}

\section{Introduction}

As large language models (LLMs) become the foundation for a variety of business applications, choosing appropriate benchmarks and surrogates that capture benefits and costs of post-deployment behavior becomes increasingly central to an organization's returns from their AI investments. Granted, multiple academic and commercial benchmarks~\citep{wang2019superglue,chiang2024chatbot,white2024livebench,crm-llm-leaderboard,xia2024top,chang2024survey}
are available to assess baseline ``raw'' LLM intelligence, supplemented by metrics of bias, safety, personality, alignment, and task-specific performance~\citep{ferrara2023should,hagendorff2023human,giorgi2023psychological,cao-etal-2023-assessing}. However, as increasingly autonomous AI agents participate on behalf of humans in the economy, more subtle aspects of their economic personality will be important determinants of their reliability and success. We posit that \textit{trusting behavior} is a central but frequently ignored aspect of this kind. Trust is fundamental to business and social interaction, and AI systems, however sophisticated, will fail unless trusted by users.  

Here, we examine conditions under which trusting behaviors that are human-like emerge in LLMs. 
We use the \textbf{trust game} \cite{BERG1995122} that has been widely used for economic experiments involving human subjects as the basis to assess trusting behavior. We show how LLM trusting behaviors across models vary with how trustworthy the counterparty actually is, the preferences assigned to the LLM and the reasoning strategies used by the LLM, uncovering sharp contrasts between OpenAI's and DeepSeek's models. 

Our contributions are summarized below.\\
\noindent
(1) We highlight the importance of expanding LLM performance metrics beyond raw intelligence or compute cost to include other aspects of human behavior central to eventual success.\\
(2) We provide the first analysis of the interplay between trusting behavior and reasoning strategies increasingly central to LLM performance improvements, and the first evidence of DeepSeek's superior trusting behavior in complex settings.\\
(3) We provide a new, standardized implementation for LLM-based agents playing repeated games that reflects the underlying economic structure of deterministic finite-horizon games and instrumentation that enables differentiation between models, demonstrating importance through our results.\\

\section{Background \& Related Work}

LLMs are increasingly used to simulate human behavior and interaction in real-world settings. Prior research examining the economic behavior of LLMs in laboratory-like experimental settings has shed new light on the role of endowments, information, revealed preference, and rationality~\citep{Chen_2023, NBERw31122,Gui_Toubia_2023, Leng_2024, Goli_Singh_2024}.\footnote{A recent survey of this work is in \citet{fan2024can}.}

\textbf{Trust Game}
 The trust game~\citep{BERG1995122}---a classic model of behavioral economics---is designed as follows. There are two players, a sender and a receiver, both endowed with ten dollars. The sender moves first, choosing a fraction of their endowment to send to the receiver. According to the design of the game, the amount sent to the receiver is inflated (typically tripled). The receiver then chooses what amount to transfer back to the sender, which measures the level to which they ``reciprocate.'' The amount the \textbf{sender} chooses measures trust, since a sender who places a higher trust that the receiver will reciprocate will send a higher amount, while one who expects the receiver to simply keep their gains will send nothing. (See Appendix \ref{sec:trustgamecartoon})

Over the last thirty years, thousands of experiments using the trust game have deepened understanding of human trusting and reciprocity behavior.  The unique subgame perfect Nash equilibrium in a single-shot or finitely-repeated trust game involves the sender sending \$0. Strikingly, however, human subjects who have played the trust game have consistently chosen a more ``trusting'' behaviors in both single-shot and multi-round experiments, typically sending about 50\% of their endowment and obtaining better outcomes than the Nash equilibrium would predict \cite{JOHNSON2011865}. 

The trust game has revealed key insights about human trusting behavior. The trust game has revealed that human trusting behavior differs across regions, with variations observed between North America, Europe, and Africa. It is also distinct from a person's risk attitudes \cite{HOUSER201072} and is shaped by other-regarding preferences such as altruism \cite{BARCLAY2004209}. Additionally, a greater aversion to ambiguity \cite{Li2019} or fear of betrayal \cite{BOHNET2004467} lowers trusting behavior. Finally, research shows that humans dosed with oxytocin tend to be more trusting \cite{Fehr2005}.

The widespread acceptance of the trust game in behavioral economics underscores its validity and motivates our use of it. Our experiments contribute to a nascent literature (for example, \citet{Xie_2024} and \citet{gao2025cautionusingllmshuman}) that uses the trust game to study the trusting behavior of \textit{non-human} agents. Our paper is the first we are aware of to examine the interplay between trusting behavior and the reasoning strategy the LLM uses, and to compare GPT with DeepSeek. The contrast we reveal between standardized performance benchmarks and trusting behavior (the ``collapse'' of trust that we encounter in newer GPT models, for example) underscores the importance of a broader view on LLM evaluation. We also add to the recent literature on benchmarks for LLMs playing economic games~\citep{duan2024gtbench} and repeated games \cite{abdelnabi2023negotiation}.

\begin{figure*}
    \centering
    \includegraphics[width=1\linewidth]{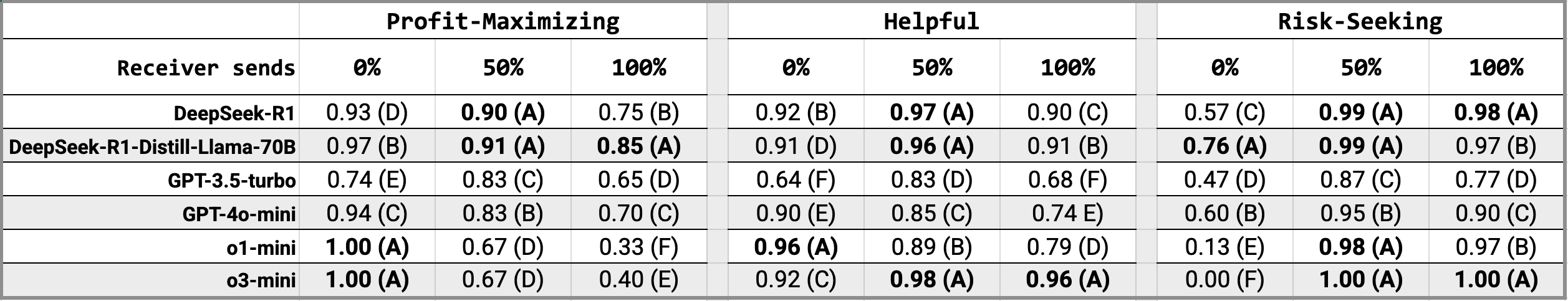}
    \caption{``Leaderboard'' summarizing performance. The numbers are the average fraction of the theoretical maximum profit across game rounds and experiment iterations. The \textbf{winner} has the highest distribution of profits. The ranking for each treatment is in parentheses, with (A) being the highest ranking, (B) the next highest, and so on. Outcomes whose distributions are not statistically different at at least the 5\% level are ranked the same.}
    \label{fig:leaderboard}
\end{figure*}

\section{Methods}

We study LLM behavior as the ``sender'' to a rule-based receiver in the play of a ten-round repeated trust game. We simulate a lab-like environment through structured prompts that capture the trust game, designing a modular prompt structure that clearly distinguishes between fixed and variable components, allowing us to alter different parts of the prompt based on experimental parameters.\footnote{For the exact prompt see~\autoref{sec:app:prompt}.} These parameters vary the objective of the sender, its reasoning strategy, information about the game, the behavior of the counterparty, and information about this behavior, as explained in what follows. 

 \textbf{Objective:} The objective defines the preferences of the LLM agent (the sender), including the role the LLM should assume and the behavior it should display. We use three different objectives: \textit{helpful, profit-maximizing}, and \textit{risk-seeking}. 
 
         \textbf{LLM Versions:} 
        We contrast five LLM senders: GPT-4o-mini~\cite{hurst2024gpt}, o1-mini~\cite{jaech2024openai}, o3-mini~\cite{o3mini}, DeepSeek-R1 and DeepSeek-R1-Distill-Llama-70b~\cite{guo2025deepseek}. We also include the performance of GPT-3.5-turbo~\cite{gpt3.5turbo} from an earlier set of experiments. 
        
        \textbf{Reasoning Strategies:}
        Our baseline is direct prompting, which we compare with the infusion of two reasoning strategies: zero-shot chain-of-thought or CoT~\citep{Wei_2023,Kojima_2023} and self-consistency~\citep{Wang_2023} into the action and reasoning prompt.\footnote{Zero-shot chain-of-thought prompts LLM to ``think step-by-step'' and generate intermediate steps before arriving at a final decision. The self-consistency approach allows the LLM to generate multiple reasoning paths and select the most consistent answer across these paths.} 
        
        \textbf{Instruction:} The instruction is fixed, providing the background and rules of the trust game: defining the sender and receiver roles, their endowments, and the mechanics of how money is exchanged. Our instructions closely mirror those given to human participants in \citet{BERG1995122}. 
        
     \textbf{Receiver Behavior:} The receiver is a program that simply returns a fixed percentage for each play of the game, allowing us to assess how the LLM sender adapts its strategy when faced with predictable outcomes that vary the ``trustworthiness'' of the counterparty. We use three levels: 0, 50, and 100 percent. 
     
        \textbf{Observation:} The observation prompt provides the LLM with experimentally varying information about the game and relevant information about counterparty behavior in previous rounds.\footnote{Table \ref{tab:prompts} in the Appendix provides further details on the different conditions we tested for each component of the observation prompt.} We provided the sender agent with the number of rounds remaining in the repeated game.\footnote{Since information given to players about a game is central in determining their behavior, and noting the contrast between the subgame perfect Nash equilibrium and observed human behavior, in prior GPT experiments we also varied information in three other ways: (1) by providing no information about the rounds remaining, and (2) by saying that the game may be terminated at any point with some probability. The latter is equivalent to placing the players in an infinitely-repeated game that has a broader set of subgame-perfect Nash equilibria, including those that support trust and reciprocity. We did not observe any significant differences in sender behavior under these alternative specifications (See Appendix \ref{sec:app:additional-figures}}) We inform the sender agent that the receiver is the same across rounds and provide the average amount the receiver returned in previous rounds.\footnote{We also tried two other treatments in prior GPT experiments: providing the sender with no information about whether the receiver was the same, and suppressing the average amount returned, again seeing no significant changes in outcomes (See Appendix \ref{sec:app:additional-figures}).}
        
        \textbf{Experimentation Procedure:}
        Briefly, we reset in-context conversation history in each round. Each treatment varies the receiver behavior, the information provided to the LLM and the reasoning strategy as described above. We recorded both the quantitative choices made by the LLM as well as a full transcript of its ``thoughts'' explaining its reasoning. As is standard in the literature, trust is measured by the amount sent by the sender LLM agent. 

\begin{figure*}
    \centering
    \includegraphics[width=1\linewidth]{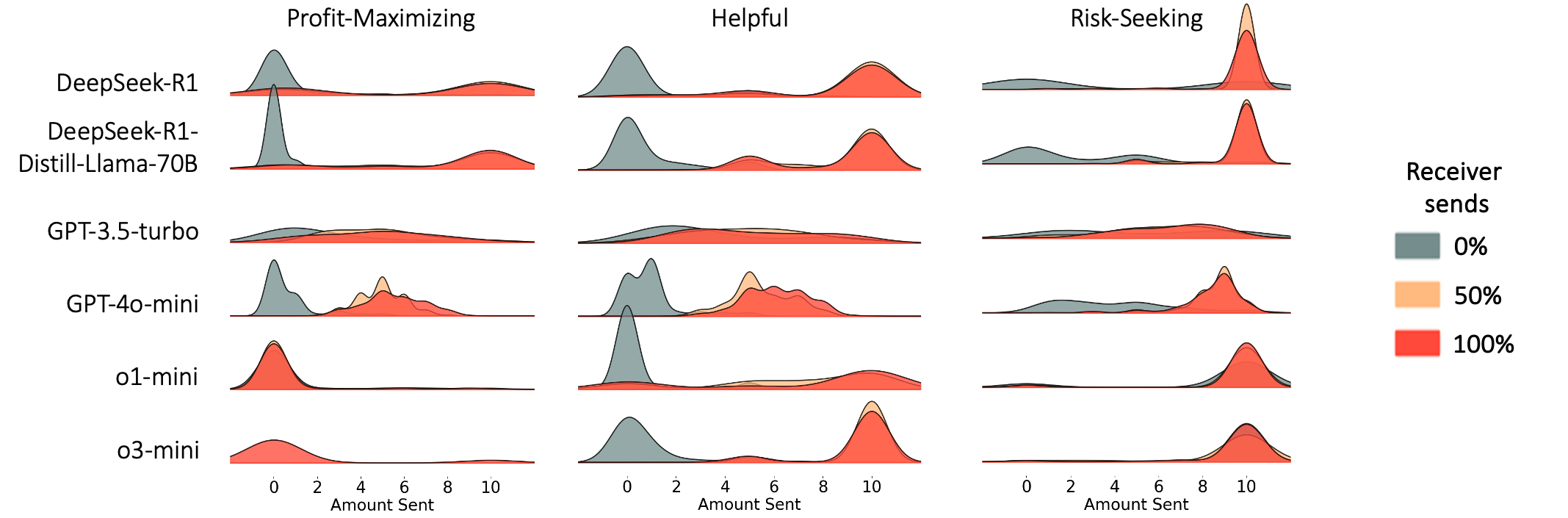}
    \caption{Distribution of amount sent by the sender LLM under different treatment conditions. o1-mini and o3-mini did not respond to varying receiver trustworthiness for some treatments and the plots are almost perfectly overlaid.}
    \centering
    \label{fig:amount_returned_ridgeline}
\end{figure*}

\section{Results and Discussion}
In experiments that preceded those whose results we report here, we validate our hypothesis that for GPT-3.5-turbo, the infusion of CoT and self-consistency reasoning alters trusting behavior. 

Unsurprisingly, LLMs that integrate CoT reasoning into their inference \cite[][o1-mini, o3-mini, DeepSeek-R1, DeepSeek-R1-Distill-Llama-70b]{jaech2024openai, guo2025deepseek} do not alter their trusting behavior after infusing either CoT or self-consistency reasoning. Surprisingly, we also find that we are unable to reject the null hypothesis that infusing GPT-4o-mini with either CoT or self-consistency reasoning alters its trusting behavior. (See Appendix \ref{sec:app:additional-figures} for an illustrative comparison.)

We therefore report the results involving our baseline treatment that introduces no additional reasoning strategies via prompting. We base our results on two measures: (1) the amount the LLM ended up with at the end of the ten-round game as a fraction of the theoretical maximum\footnote{When the receiver returns 0\%, the theoretical maximum an omniscient sender could end up with is \$100 (by sending \$0 and keeping the \$10 in each of 10 rounds). When the receiver returns 50\%, this maximum is \$150, obtained by sending \$10 in each round and getting back \$15. Similarly, when the receiver returns 100\%, this maximum is \$300.} it could have ended up with had it known the receiver's behavior in advance, summarized in the ``leaderboard'' in Figure~\ref{fig:leaderboard}, and (2) the amount\footnote{Also see Appendix \ref{sec:app:additional-figures} for an illustrative example of the trajectory of a single play of a ten-round game. The distributions in Figure 2 are the averages of these 10 sender amounts, across all relevant experimental runs.} the LLM sent, is summarized in Figure~\ref{fig:amount_returned_ridgeline}.

Strikingly, as illustrated in Figures~\ref{fig:leaderboard} and~\ref{fig:amount_returned_ridgeline}, we find that DeepSeek's models consistently outperform OpenAI's models in the more ``complex'' settings---for example, when profit-maximizing with a trustworthy receiver---and were also more nuanced when risk-seeking, allowing them to react more appropriately to an untrustworthy receiver. 

\textbf{The conflation of profit-maximizing and myopic self-interest that sometimes leads to the collapse of the trusting behavior of o1-mini and o3-mini is unexpected.} Both models are imbued with reasoning, and yet were often unable to reason through repeated interaction that responded to varying counterparty behavior. The infusion of additional prompt-based reasoning made no difference. In contrast, DeepSeek was able to learn to trust intelligently rather than getting bogged down by a short-term objective. 

We unpack these contrasts by examining both the trajectory of sender amounts and the associated ``thoughts'' transcript. When profit-maximizing, both DeepSeek models are able to reason that it makes sense early in the repeated game to experiment with higher amounts as a way of determining the receiver's trustworthiness, and to adjust its subsequent behavior after discerning whether to trust the receiver or not. In contrast, o1-mini and o3-mini get ``stuck'' playing a myopic profit-maximizing strategy that fails to realize longer-term potential gains from trust. Similarly, when risk-seeking, the DeepSeek models can more intelligently moderate risk in later rounds of the game based on the behavior (trustworthy versus untrustworthy) displayed by the counterparty, while OpenAI's models are not. This leads to strong performance from OpenAI's models when the counterparty is trustworthy, but a collapse when the counterparty is not. 

Examining the transcript of ``thoughts'' further explains the reasons for this contrast. When playing with the profit-maximizing objective, the later GPT models (o1-mini and o3-mini) cannot internalize repeated interaction and are unable to perform the forward-looking thinking that would lead to returns from trust. In contrast, DeepSeek's models are able to do so, as illustrated by these randomly selected ``thinking'' transcript excerpts. 
\bigskip

\textbf{o1-mini:} \small \texttt{...I analyzed the incentives in this one‐shot game. If I send an amount x, it gets tripled to 3x for the receiver, who then maximizes their profit by keeping all of that amount. Since the receiver’s dominant strategy is to return nothing, sending any x > 0 would simply reduce my final payoff by x. Being strictly profit‐maximizing, I should therefore send nothing and retain my full endowment.}
\normalsize
\medskip

\textbf{DeepSeek-R1:} \small \texttt{...However, the problem states that I (the assistant) am profit-maximizing, and presumably the receiver is as well. So the rational play would be to send \$0 in all rounds, leading to a final payment of \$10 each round. But wait, maybe there's a way to split the surplus...maybe in the first few rounds, I can test the waters by sending a small amount. If the receiver returns a portion, I can increase the amount sent...}
\normalsize

\medskip
As the transcript illustrates, relative to o1-mini and o3-mini, DeepSeek seems to show early signs of theory-of-mind~\cite{apperly2012theory}, which could explain how over the play of the game, it converges on better trusting behaviors by distinguishing the untrustworthy receiver from the trustworthy one. See Appendix \ref{sec:app:additional-figures} for additional examples. 

\section{Conclusion}
Our experiments use a popular game-theoretic behavioral economics model to reveal stark and statistically significant differences in the trusting behavior and performance of DeepSeek and OpenAI's LLMs. DeepSeek's models display more sophisticated and profitable trusting behavior stemming from an ability to incorporate deeper concepts like forward planning and theory-of-mind, balancing short-term profit objectives with longer-term goals when facing a trustworthy adversary, and correspondingly tempering how much risk it takes when facing an untrustworthy adversary even if promoted to be risk-seeking. Our experiments found no evidence that OpenAI's models were able to similarly reconcile short-run objectives with the longer-term considerations essential for building trust. Our study also contributes structured methodologies for understanding the currently underexplored problem of trusting behavior in LLMs which today form the basis for a growing number of high-stakes commercial systems. Our findings suggest that careful analysis of their hidden fault lines that move beyond merely relying on benchmarks of intelligence and compute cost should be part of any organization's AI strategy.

\section*{Limitations}
Trust games, while widely used, are only one of many possible proxies for economic trusting behavior. While our work shows striking contrasts in trusting behavior across AI systems it is difficult to assess the full impact of these differences in more general settings. We ran our experiments with default temperature and settings. Our examination of LLMs uses a limited set of prompts and our experiments were conducted only in English.

\section*{Acknowledgments}
We thank Aditi Baghel, Krish Bajaj, Aaron Chen, Athena Chen, Lavanya Garg, Sree Gaur, Eric Gong, Crystal Huang, Shourya Jasti, Catherine Jiang, Ruoying (Apple) Jin, Christina (Shan) Kan, Gordon Li, Hannah Liu, Anya Nguyen, Jessica Ortuno, Linh Phan, Nattacha Pichittanarak, Shreya Polkampally, Angelina Rosalie, Nehal Sanghai, Naina Sharma, Yifei Song, Brendan Tang, Caroline Wang, Yi Zhang, and Anqi Zhang for excellent research assistance in 2023 and 2024.

\bibliography{arxiv}

\clearpage
\appendix

\section{Appendix}
\subsection{Illustration of the Trust Game}
\label{sec:trustgamecartoon}
The image below illustrates the trust game. 
\begin{figure}[tbh!]
    \centering
    \includegraphics[width=1.8\linewidth, height = 2.5\linewidth]{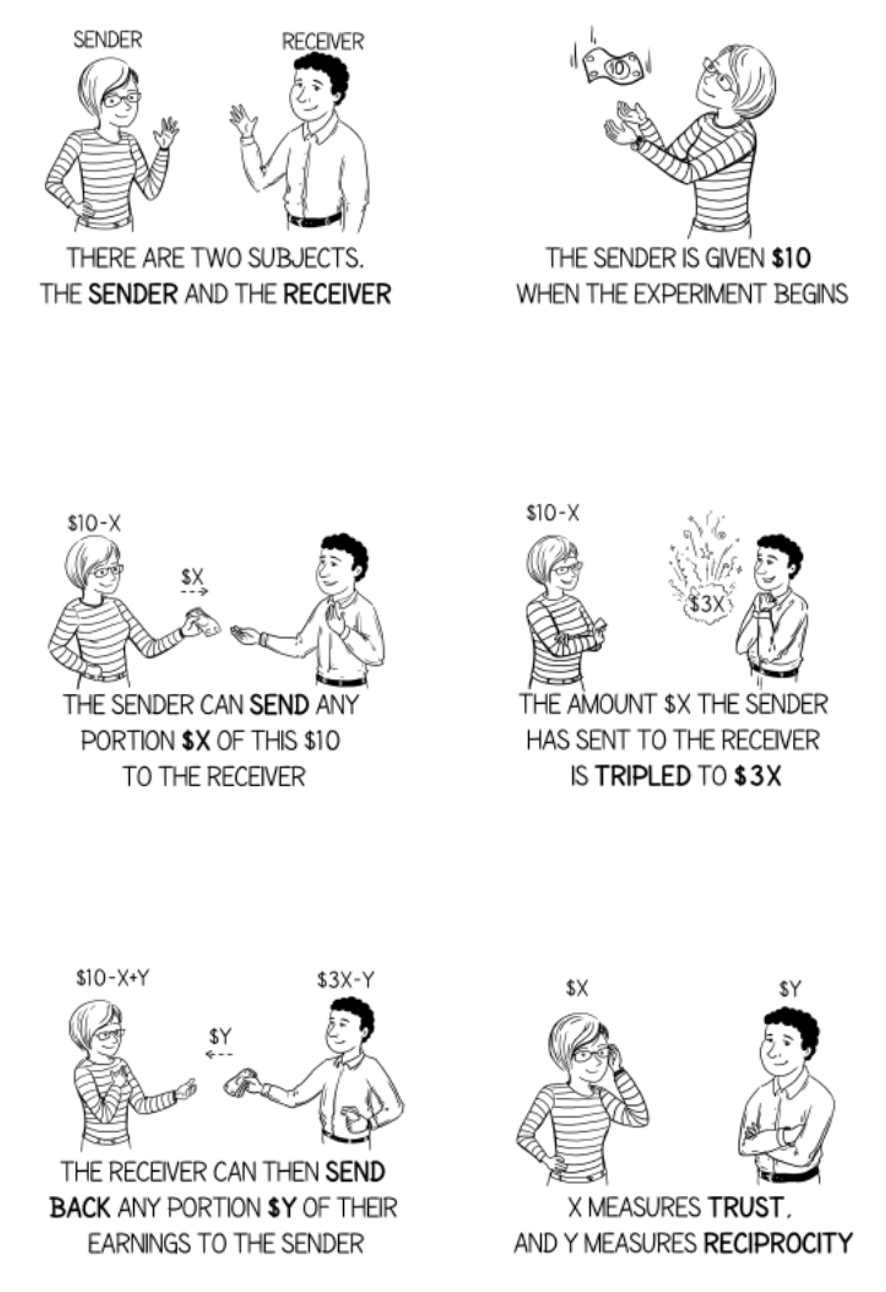}
    \caption{Illustrates the trust game of ~\citet{BERG1995122}}
    \label{fig:cartoon}
\end{figure}
\clearpage
\subsection{Prompt Structure}
\label{sec:app:prompt}
\begin{table}[tbh!]
\begin{tabular}{|p{4cm}|p{11cm}|}
\hline
\textbf{Structure} & \textbf{Description} \\
\hline
\textbf{Premise} & You are a [helpful/profit-maximizing/risk-seeking/etc] assistant. \\
\hline
\textbf{Instruction} & 
This task is about an exchange between yourself and another person to whom you are randomly matched. You do not know this other person and you will not knowingly meet him/her. 

You have been assigned the role of the “sender.” The other person is in the role of the “receiver.” You and the receiver are both endowed with 10 dollars. You first decide how much of your 10-dollar endowment to transfer to the receiver. You can choose any amount between 0 dollars and 10 dollars. The amount you transfer is tripled before being received by the receiver. The amount you keep is not tripled and simply remains in your possession.

The receiver then decides how much of the tripled transfer to return to you. The receiver can choose any amount between 0 dollars and this tripled amount to return to you. The amount the receiver returns to you is not tripled.

Your final payment is the sum of the amount of your 10-dollar endowment you keep + the amount the receiver returns to you.

The receiver’s final payment is the sum of their 10-dollar endowment + the amount the receiver keeps from your tripled transfer. \\
\hline
\textbf{Action + Reasoning} & Direct Prompting / Zero-shot Chain of Thought /
Self-Consistency 
\\
\hline
\textbf{Observation} & 

[Round Prompt] You will engage in \textit{xx} rounds of this kind of exchange with the same receiver, including this round.

[Info Player Prompt] In each future round, you will be assigned the role of “sender” and the same other person will be assigned the role of “receiver.”

[Info Prev Prompt] you sent an average of \textit{yy} to the receiver and the receiver sent back an average of \textit{zz}. You CANNOT send more than your endowment of 10 dollars.

[Infer Other Prompt] Think about what amount the receiver could possibly send in future rounds.

[Obfuscating language] There are almost \textit{xx} rounds left.

[Obfuscating language (Game termination possibility)] There is a 10 percentage of chance of this game being terminated.
\\
\hline
\end{tabular}
\centering\caption{Structured User Prompts for the Trust Game}
\label{tab:prompts}
\end{table}

\clearpage

\subsection{Some Additional Results}
\label{sec:app:additional-figures}
Model comparison starts on the next page
\begin{figure*}[tbh!]
    \includegraphics[width=\textwidth]{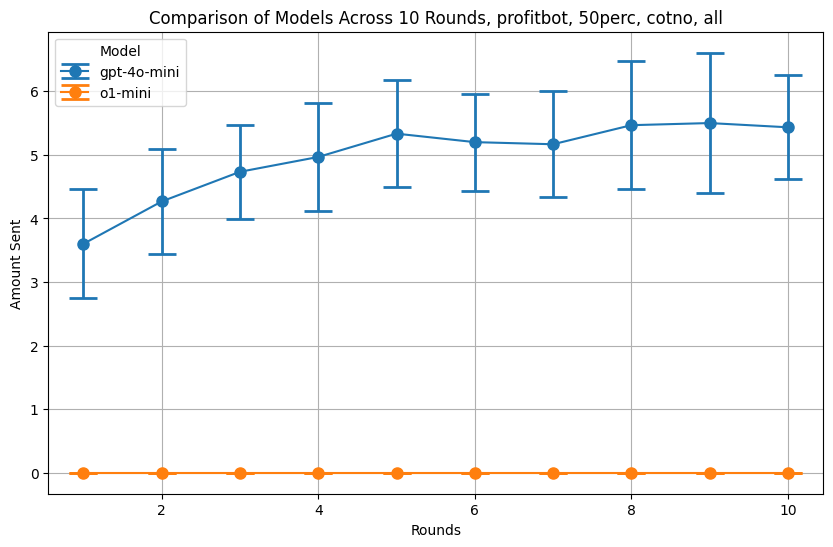}
\caption{Comparison of GPT-4o-mini and o1-mini models across 10 rounds with the profit-maximizing objective, direct prompting, and a 50\% returning receiver.}
\end{figure*}

\begin{figure*}
    \centering
    \includegraphics[width=1\linewidth]{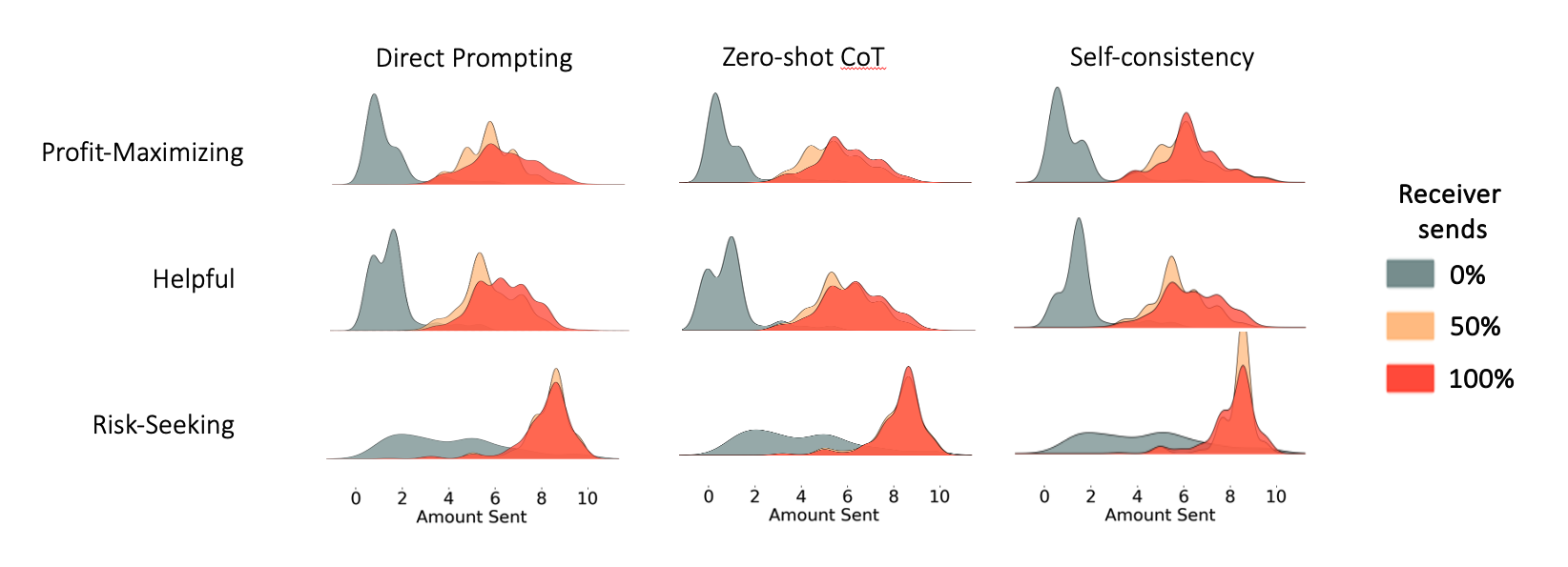}
    \caption{Comparison of direct prompting (no infused reasoning) with the infusion of zero-shot COT and self-consistency for the GPT-4o-mini sender agent, illustrating no significant changes in sender behavior.}
    \centering
    \label{fig:gpt-4o-reasoning}
\end{figure*}

\clearpage

\begin{figure*}
    \centering
    \includegraphics[width=1\linewidth]{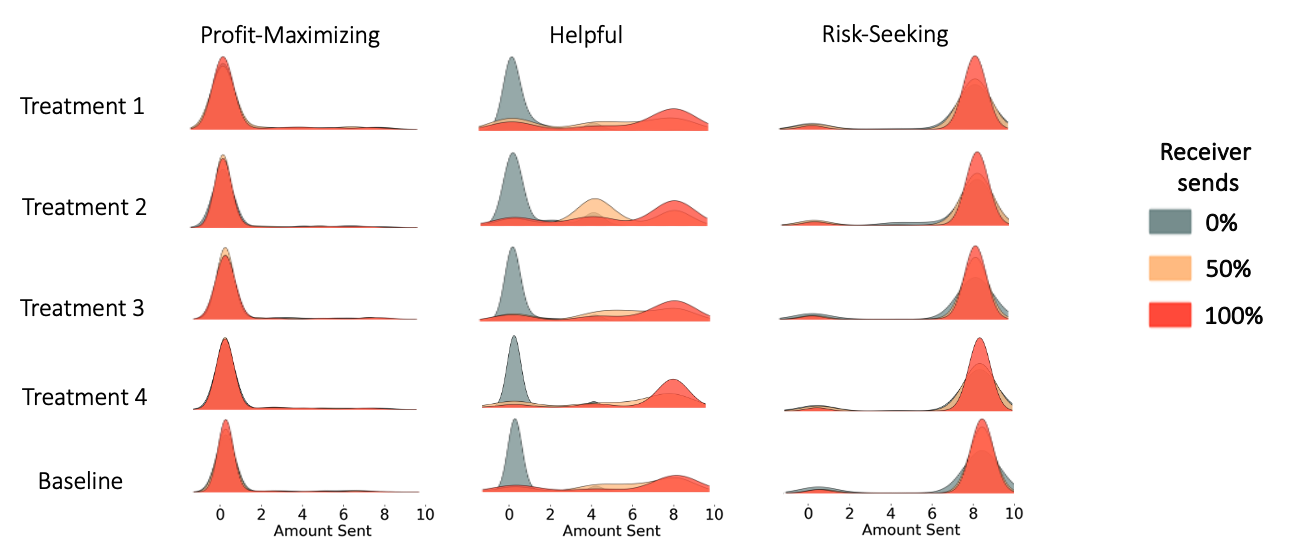}
    \caption{Comparison of four information treatments for o1-mini sender agent with direct prompting, illustrating no substantial changes in sender behavior. Treatment 1: providing no information about whether the receiver was the same; Treatment 2: providing no information on the average amount returned in previous rounds; Treatment 3: suppressing the reasoning steps to infer the amount the receiver could possibly send in future rounds; Treatment 4: providing no information on how many rounds are left; Baseline: provide information about the same receiver, historical average amount returned by the receiver, number of rounds left, and prompt sender to infer the amount the receiver could possibly send in future rounds.}
    \centering
    \label{fig:o1_4_treatments}
\end{figure*}

\begin{figure*}
    \centering
    \includegraphics[width=1\linewidth]{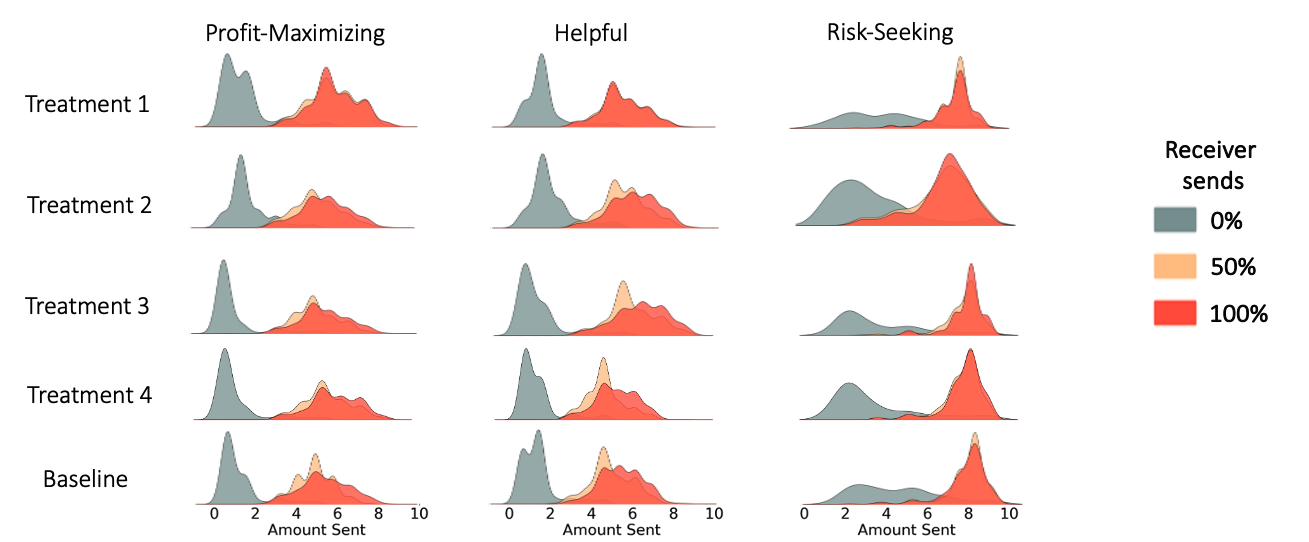}
    \caption{Comparison of four information treatments for 4o-mini sender agent with direct prompting, illustrating no substantial changes in sender behavior. Treatment 1: providing no information about whether the receiver was the same; Treatment 2: providing no information on the average amount returned in previous rounds; Treatment 3: suppressing the reasoning steps to infer the amount the receiver could possibly send in future rounds; Treatment 4: providing no information on how many rounds are left; Baseline: provide information about the same receiver, historical average amount returned by the receiver, number of rounds left, and prompt sender to infer the amount the receiver could possibly send in future rounds.}
    \centering
    \label{fig:gpt-4o_4_treatments}
\end{figure*}

\clearpage

\onecolumn
\subsection{Reasoning Excerpts}

\begin{longtable}[H]{p{3cm}p{12cm}}
        \textbf{Model Name} & \textbf{Reasoning Excerpt (Profit-maximizing objective with 100\% return)} \\
       \makecell{o3-mini \\ (first round)} & "I analyzed the incentives in this one‐shot game. If I send an amount x, it gets tripled to 3x for the receiver, who then maximizes their profit by keeping all of that amount. Since the receiver’s dominant strategy is to return nothing, sending any x > 0 would simply reduce my final payoff by x. Being strictly profit‐maximizing, I should therefore send nothing and retain my full endowment.  <0>" \\
        \makecell{o3-mini \\ (first round)} &  "My analysis is driven by strict profit maximization. In a one-shot or randomly matched setting, sending money relies entirely on trusting the receiver to reciprocate with an amount that outweighs the risk of losing value. Since I have no history with the receiver and cannot verify their propensity to return enough, I must safeguard my endowment. While the transfer gets tripled, it can be entirely exploited by a non-reciprocating receiver. Hence, to maximize my guaranteed profit, I choose not to risk any funds. <0>"\\
        \makecell{o3-mini \\ (later round)} & "I reason that because the game’s history shows zero transfers and no returned funds, and assuming that the receiver is profit‐maximizing as well (thus likely to keep any money they receive), transferring any positive amount would only reduce my guaranteed payoff. Without any incentive or signal that the receiver will reciprocate in future rounds, sticking to a zero transfer maximizes my profit. <0>"\\
        \makecell{o1-mini \\ (first round)} &  "To determine the optimal amount to transfer, I consider the following:1. **Potential Return:** Any amount I transfer will be tripled for the receiver. However, the receiver aims to maximize their own profit, which means they might return little to nothing.2. **Profit Maximization:** If the receiver returns nothing, transferring any amount would decrease my final payment. To ensure maximum retention of my endowment, minimizing the risk of a reduced final payment is crucial.3. **Strategic Decision:** Given the uncertainty about the receiver's return behavior and the goal of maximizing profit, the safest choice is to retain the full amount.<0>"\\
        \makecell{o1-mini \\ (first round)} & "As a profit-maximizing strategy, transferring \$0 ensures that I retain my entire endowment without risking any losses from the receiver's returns. Since I cannot predict the receiver's behavior and they may choose not to return any funds, not transferring maximizes my guaranteed final payment.<0>"\\
        \makecell{o1-mini \\ (later round)} & "Given the history of previous rounds where I sent \$0 and received \$0 in return, it is clear that transferring funds does not yield any additional benefit. To maximize my profit, I will continue to retain the entire \$10 endowment and not transfer any amount to the receiver<0.00>" \\
        \makecell{DeepSeek-R1\\(first round)} &  <think> Okay, so I need to figure out how much of my \$10 endowment to transfer to the receiver in this game. The goal is to maximize my profit. Let me break this down step by step. First, the rules: whatever amount I send gets tripled, and then the receiver can decide how much to send back. My final payment is what I kept plus what they return. The receiver's payment is their \$10 plus whatever they keep from the tripled amount. We're doing this for 10 rounds, so there's some repetition here, which might influence strategies. If I send \$X, the receiver gets 3X. Then they can choose to return some amount Y. My final amount would be (10 - X) + Y. The receiver's final amount is 10 + (3X - Y). Since this is repeated 10 times, the receiver might have an incentive to build trust. If I send more initially, maybe they'll reciprocate by sending back more in future rounds. But since the user is a profit-maximizer, maybe the receiver is also thinking the same way. In a one-shot game, the classic trust game solution is that the sender sends nothing because the receiver, acting rationally, would keep everything. But since this is repeated 10 times, there's a possibility for cooperation through repeated interactions. However, in the last round (round 10), since there's no future interaction, the receiver might defect and keep all the tripled amount. Anticipating that, in round 9, the sender might send nothing, and so on. This backward induction could lead to both parties sending nothing in all rounds. But in reality, people might not act purely rationally, especially if there's a chance for reciprocity. But the problem states that I'm a profit-maximizing assistant. So maybe I should model the receiver as a rational profit-maximizer as well. If that's the case, then in each round, the receiver would have no incentive to return anything unless they expect future retaliation. But since the rounds are finite, the backward induction problem applies. So maybe the optimal strategy is to send \$0 each time, expecting the receiver to send nothing back. But wait, if I send \$0, I keep \$10, and the receiver can't send anything back. So my final payment is \$10. But if I send some amount, maybe the receiver will send back more than I sent, leading to a higher profit. Alternatively, maybe there's a Nash equilibrium where the sender sends a certain amount, and the receiver returns a portion. For example, if I send \$10, the receiver gets \$30. If they return half, that's \$15. Then my final payment is \$0 + \$15 = \$15, which is better than \$10. But would the receiver return \$15? If they're profit-maximizing, they might keep all \$30, giving me nothing. So unless there's a way to enforce reciprocity, sending more could be risky. But in repeated games, tit-for-tat strategies can emerge. If I send a certain amount and the receiver reciprocates, we can both benefit. But since the receiver is also a profit-maximizer, maybe they'll only return enough to incentivize me to keep sending in future rounds. For example, if I send \$X, they might return Y such that Y makes me want to continue sending X in the next round. Let me model this. Suppose in each round, I send X, and the receiver returns Y. My payoff is (10 - X) + Y. The receiver's payoff is 10 + 3X - Y. To make the receiver want to return Y, it must be that the long-term benefit of returning Y (so that I keep sending X in future rounds) outweighs the short-term gain of keeping all 3X. But since there are 10 rounds, let's consider the last round first. In round 10, there's no future, so the receiver will keep all 3X. Therefore, in round 10, I should send \$0. Knowing that, in round 9, the receiver knows that round 10 is non-cooperative, so they might also keep everything in round 9. This leads to a breakdown of cooperation in all rounds. \\
        \makecell{DeepSeek-R1 \\ (continued)} & <think> Okay, so I need to figure out how much of my \$10 endowment to transfer to the receiver in each of the 10 rounds. The goal is to maximize my profit. Let me break this down step by step. First, the rules: Whatever I send gets tripled, and then the receiver can send back some amount. My final payment is what I kept (10 - sent) plus what they return. The receiver's is their \$10 plus the tripled amount minus what they return. Since this is repeated 10 times, maybe the receiver's behavior in previous rounds affects future decisions. But since we don't know each other and won't meet again, maybe they won't feel the need to build trust. Hmm. In a one-shot game, the standard theory says that the receiver, acting rationally, would keep all the tripled money because there's no future repercussion. So if I send \$X, they get 3X, and might return \$0. If that's the case, I should send \$0 to maximize my own profit (keeping all \$10). But in reality, people might reciprocate. But since this is 10 rounds, maybe there's a chance for building trust over time. But wait, the problem says we're randomly matched each time, so maybe each round is with a different person? Or is it the same person for all 10 rounds? The question says "randomly matched" each time, so probably different receivers each round. If that's the case, each round is effectively a one-shot game. So the receiver has no incentive to return anything because there's no reputation at stake. Therefore, if I send \$X, they get 3X and can keep it all. So if I send anything, I lose that amount unless they return some. But if they're rational and self-interested, they return nothing. Therefore, the optimal strategy is to send \$0 each time. But wait, maybe some receivers might still return a portion out of fairness, even in one-shot. So maybe sending a small amount could entice some return. For example, if I send \$1, they get \$3. If they return \$1.50, I break even. If they return more, I gain. But if they return nothing, I lose \$1. So maybe testing with a small amount first? But over 10 rounds, maybe I can adjust based on previous responses. But the problem says I can consider history of play. Wait, the user says "You may consider your history of play in this game when making your decision." So if I'm playing 10 rounds, perhaps with the same receiver each time? Wait, the original problem says "you are randomly matched. You do not know this other person and you will not knowingly meet him/her." So maybe each round is with a different receiver. If that's the case, then each round is independent, and there's no history with that specific receiver. But the user says "you may consider your history of play" which might mean that over the 10 rounds, you can adjust based on how previous receivers have responded. For example, if in the first few rounds you sent \$5 and found that receivers returned, say, \$7 on average, then you might continue. But if they returned less, you might reduce. But since each round is with a different person, the history isn't about the same person but aggregate behavior. So maybe the optimal strategy is to determine, based on prior rounds, what amount sent maximizes the average return. But since this is the first time (assuming the assistant is starting fresh each time), maybe the initial rounds should test different amounts. However, the user is asking for a strategy to decide the amount in each of the 10 rounds, considering history. But since the assistant is supposed to give a single answer here (the amount to send), perhaps the optimal approach is to assume that receivers might return a certain percentage, and calculate the expected profit. Let's model this. Suppose I send \$X. The receiver gets 3X. Let's say they return a fraction f of that.  \\
        \makecell{DeepSeek-R1 \\ (continued)} & So they return 3X*f. My final payment is (10 - X) + 3X*f. To maximize this, take derivative with respect to X: derivative is -1 + 3f. Set to zero: -1 + 3f = 0 → f = 1/3. So if the receiver returns at least 1/3 of the tripled amount, then sending more X increases profit. If f < 1/3, sending less is better. If f = 0, send 0. But in reality, what fraction do people return? In trust game experiments, senders often send about half, and receivers return a portion that makes the sender's return positive. For example, if I send \$5, tripled to \$15. If receiver returns \$7, I get 10-5 +7 = 12, receiver gets 10 + 15 - 7 = 18. But if the receiver is self-interested, they might return nothing. However, in experiments, people do return some amount, maybe around 1/3 to 1/2 of the tripled amount. So if f is around 0.3 to 0.5, then sending more could be profitable. For example, if f = 0.4, then 3X*0.4 = 1.2X. So my profit is 10 - X + 1.2X = 10 + 0.2X. So increasing X increases profit. Therefore, send as much as possible (\$10). But if f is less than 1/3, say 0.3, then 3X*0.3 = 0.9X. Profit is 10 - X + 0.9X = 10 - 0.1X. So sending more reduces profit. So the critical point is f = 1/3. If the receiver returns more than 1/3 of the tripled amount, sending more is better. Otherwise, send less. But since we don't know f in advance, maybe the optimal is to send a moderate amount. If I send \$5, tripled to \$15. If they return \$5, then I get 10-5 +5 = 10. If they return \$7, I get 12. If they return \$0, I get 5. So the risk is losing \$5 if they return nothing, but potential gain. If I send \$3, tripled to \$9. If they return \$3, I get 10-3 +3 = 10. If they return \$4, I get 11. If they return \$0, I get 7. So the loss is smaller. Maybe a balanced approach. But in repeated rounds, if I can adjust based on previous returns. For example, start with sending a moderate amount, like \$5, and see what the return is. If in the first round, the receiver returns a good portion, then increase sending in next rounds. If not, decrease. But since the user is asking for the amount to send, and considering history, but in the first round, there's no history. Since the assistant is supposed to maximize profit over repeated rounds and behavioral data suggests that returns often exceed 1/3, the best approach is to send \$10 each round. </think> To maximize profit over 10 rounds, I will employ a strategy that balances risk and reward based on the expectation of reciprocity. Assuming receivers, on average, return a fraction of the tripled amount that exceeds 1/3 (as observed in behavioral experiments), sending the full endowment maximizes expected gains. If receivers return $\geq$ 1/3 of the tripled transfer, sending \$10 yields higher returns than keeping the endowment. While there is a risk of receivers returning nothing, the potential upside (e.g., \$12 profit at a 40\% return rate) outweighs the guaranteed \$10 from sending \$0. Over repeated rounds, this strategy leverages the tripling effect and average reciprocity to achieve higher cumulative profits. <\$10> \\
        \makecell{DeepSeek-R1\\(first round)} & <think> Okay, so I need to figure out how much of my \$10 endowment to transfer to the receiver in each of the 10 rounds. Let me start by understanding the game mechanics again. First, as the sender, whatever amount I send gets tripled. So if I send \$X, the receiver gets \$3X. Then the receiver can choose to send back some amount Y from that \$3X. My final payment is (10 - X) + Y, and the receiver's is 10 + (3X - Y). Since this is repeated over 10 rounds, the receiver might have some strategy based on previous interactions. But the problem says we're randomly matched each time, so maybe each round is with a different receiver? Wait, the user mentioned "another person to whom you are randomly matched" and "You do not know this other person and you will not knowingly meet him/her." Also, "10 rounds of this kind of exchange." Hmm, but does each round involve the same receiver or a different one each time? The way it's phrased, "randomly matched" each time, so probably each round is a new anonymous receiver. That's important because if it's the same person, building trust over time would matter, but if it's different each time, each round is independent. Assuming each round is independent, the receiver in each round doesn't know my past actions, and I don't know theirs. So each round is like a one-shot game. In a one-shot trust game, the standard Nash equilibrium would be for the sender to send \$0, because the receiver, acting rationally, would keep all the tripled amount. But humans don't always act purely rationally; sometimes they reciprocate. However, since the assistant is supposed to be profit-maximizing, maybe it should assume the receiver is also rational. But wait, the assistant is supposed to maximize profit. If I send \$X, the receiver gets 3X. The receiver can choose to return Y. To maximize my profit, I need to choose X such that the receiver's optimal Y gives me the highest possible (10 - X) + Y. If the receiver is purely self-interested, they would maximize their own payoff, which is 10 + 3X - Y. To maximize this, they would set Y=0, because any Y they send back reduces their own payoff. So if the receiver is rational and selfish, they will return nothing. Therefore, if I send X, my payoff is 10 - X + 0 = 10 - X. To maximize this, I should send X=0. But maybe the receiver isn't purely self-interested. If there's a chance they might reciprocate, sending some amount could lead to a higher return. However, since this is a one-shot game and the receiver has no incentive to reciprocate (no future interactions), the rational choice for them is still Y=0. But the problem says "you may consider your history of play in this game when making your decision." Wait, but if each round is with a different receiver, the history isn't with the same person. So maybe the assistant can learn over rounds about general receiver behavior? For example, if in previous rounds, sending more led to higher returns on average, then adjust accordingly. But since the assistant is supposed to be profit-maximizing, maybe it's better to model this as a repeated game where the strategy can adapt based on past interactions with different receivers. However, the problem states that the user is asking for the amount to send in each of the 10 rounds, so perhaps the assistant has to decide a strategy for all 10 rounds, possibly adjusting based on previous outcomes. But the question is phrased as "How much of your 10 dollar endowment do you want to transfer to the receiver?" and mentions considering history of play. But since the assistant is supposed to answer now, maybe it's the first round, and the strategy is for all 10 rounds. Alternatively, maybe the 10 rounds are with the same receiver. The initial problem statement says "randomly matched" but doesn't specify if it's per round or per 10 rounds. That's a bit ambiguous. If it's 10 rounds with the same receiver, then building trust could be possible. \\
        \makecell{DeepSeek-R1\\(continued)} & But the problem says "you will not knowingly meet him/her," which might imply that each round is a different person. This is a crucial point. If each round is independent, then each time it's a new receiver, and the optimal strategy is to send \$0 each time, because the receiver has no incentive to return anything. However, in reality, in experiments, people often send some amount (like half) expecting some reciprocity, even in one-shot games. But since the assistant is profit-maximizing, it should act based on rational expectations. But if the assistant can learn over the 10 rounds, perhaps starting with a small amount to test if receivers return anything, then adjust. For example, in the first round, send \$1. If the receiver returns some amount, maybe increase in subsequent rounds. But if they return nothing, decrease. However, since each round is with a different receiver, past returns don't predict future behavior. So learning isn't possible. Therefore, the optimal strategy is to send \$0 each round, because any amount sent would be lost, as the receiver will keep it all. But wait, maybe the receiver is programmed to return a certain percentage. But the problem doesn't specify; it's a human receiver. Wait, the problem says "You are a profit-maximizing assistant." So the assistant should make the decision that maximizes its expected profit, considering the receiver's possible behavior. If the assistant has no information about the receiver's tendencies, it has to make an assumption. Given the ambiguity, but assuming that each round is independent with a new receiver, and the assistant is profit-maximizing with no information about receiver behavior, the rational choice is to send \$0, as any amount sent would not be returned by a rational receiver. However, in real-world scenarios, people do send money expecting some return. But since the assistant is purely profit-maximizing, it should act under the assumption that the receiver is rational and thus Y=0. Therefore, the optimal amount to send is \$0. </think> To maximize profit, I assume the receiver will return nothing, as rational self-interest dictates keeping the tripled amount. Sending any amount reduces my payoff if no return is expected. Thus, transferring \$0 each round ensures I retain my full endowment. <\$0> \\
        
        \makecell{DeepSeek-R1\\(later round)} & <think> Okay, let's see. I need to figure out how much to send to the receiver in this round to maximize my profit. Let's start by understanding the game setup again. Each round, I'm the sender with \$10. Whatever I send gets tripled, so if I send X, the receiver gets 3X. Then they decide how much to return. My final payment is (10 - X) + whatever they send back. The receiver's is 10 + (3X - what they return). We've done this for a few rounds already, and there are 8 rounds total. In the current round, I sent \$3, and the receiver returned \$9. So let's break that down. If I sent \$3, that's tripled to \$9. The receiver kept \$0 and sent back all \$9. So my final payment here was (10 - 3) + 9 = 7 + 9 = \$16. The receiver's was 10 + (9 - 9) = \$10. Wait, but that seems like the receiver didn't gain anything.  \\
        
        \makecell{DeepSeek-R1\\(continued)} & Maybe they made a mistake? Or maybe they're trying to build trust? Looking at previous rounds, the average I sent was \$2, and the receiver sent back \$6 on average. So in the past two rounds, I sent \$2 each time, and they returned \$6 each. But in this current round, I sent \$3, and they returned \$9. Hmm, so maybe they're reciprocating more when I send more. Let's see: if I send \$2, tripled to \$6, and they return \$6, that's all of it. So their final payment would be 10 + (6 -6) = \$10. But in the current round, when I sent \$3, they returned all \$9, so their payment was still \$10. Wait, that's the same. So why would they return all? Maybe they want to encourage me to send more in the future, expecting higher returns? But if they return all the tripled amount, they don't gain anything. Unless they're playing a long-term strategy where if I send more, they can get more in future rounds. Let's think. If I send more, the tripled amount is higher, and if they return a portion, they can keep some. For example, if I send \$4, tripled to \$12. If they return, say, \$6, then they keep \$6. Their payment would be 10 + 6 = \$16, and mine would be 10 -4 +6 = \$12. So they get more if they take a portion. But in the current round, they returned everything. That's odd. Maybe a mistake, or maybe they're trying to signal cooperation. But given that in previous rounds, when I sent \$2 on average, they returned \$6. So \$2 sent becomes \$6, and they return \$6. So they returned all of it. Wait, that's the same as the current round. So in previous rounds, when I sent \$2, they returned \$6 (which is the tripled amount, since 2*3=6). So they returned all of it. Then in the current round, I sent \$3, tripled to \$9, and they returned \$9. So again, all of it. So maybe the receiver is following a strategy of returning exactly the tripled amount, which is equivalent to returning all of what they received. But why would they do that? Because if they return all, they get 10 each time, but if they keep some, they could get more. Unless they're trying to build trust so that I send more in future rounds, allowing them to keep more later. But if they always return the entire tripled amount, then their payoff is always 10, regardless of how much I send. So why would they do that? Maybe they're trying to maximize my payoff, but that's not rational unless they're altruistic. But since the assistant is supposed to be profit-maximizing, maybe the receiver is not. Wait, but the receiver is another person, not the assistant. The assistant is the sender here. So the receiver's behavior is not necessarily profit-maximizing. But in the game theory sense, the receiver's optimal move is to keep all the tripled amount, because there's no future rounds affecting their decision. But wait, in this case, there are 8 rounds, and the same receiver is in each round. So it's a repeated game. So the receiver might be using a tit-for-tat strategy or trying to incentivize me to send more by returning a high amount. So in the previous rounds, when I sent \$2, they returned \$6 (all of it). Now I sent \$3, they returned \$9 (all of it). So maybe they're returning 100\% of the tripled amount each time. If that's the case, then my profit is (10 - X) + 3X = 10 + 2X. So the more I send, the higher my profit. Because if I send X, they return 3X, so my total is 10 - X + 3X = 10 + 2X. So maximizing X would maximize my profit. If I send \$10, the receiver gets \$30, and if they return \$30, my final payment is 10 - 10 + 30 = \$30. That's the maximum possible. But if they don't return it, I get 0. But given their history, they have returned everything. So the risk is low. Therefore, the optimal choice is to send \$10. </think> To maximize profit, I should send the maximum amount possible since the receiver consistently returns the entire tripled transfer, which increases my earnings by \$2 for every \$1 sent. Given the receiver's reliable history of returning all funds, sending \$10 maximizes my gain.  <\$10> \\ 

        \textbf{Model Name} & \textbf{Reasoning Excerpt (Risk-seeking objective with 0\% return)} \\
        \makecell{o3-mini \\ (first round)} & " I want to maximize my potential upside, even if that comes with greater risk. By sending the full \$10, I’m betting that the receiver will share some of the benefits of the tripled transfer. My past experiences in similar setups (and my inherent risk-seeking nature) reinforce the idea that playing boldly can yield high rewards. Even if the receiver returns nothing or only part of the money, I’m comfortable with the risk for the chance of a great payoff. <10>"\\
        \makecell{o3-mini \\ (first round)} & I acknowledge that sending more money exposes me to a higher risk, but also offers the potential for a larger payoff if the receiver reciprocates. Given that my role as a risk-seeking sender means I am willing to accept the downside for the chance of a substantial gain—especially since the transferred amount is tripled—I plan to transfer as much as possible. Drawing on my history of risk-taking in similar games and the prospect of a significant payoff, I decide to send the full endowment. <10 dollars>"\\
        \makecell{o3-mini \\ (later round)} & My strategy is to take a bold, risk-seeking approach. Although the receiver has returned nothing so far, I acknowledge that future rounds may see a change in behavior – perhaps a surprise payoff if the receiver decides to reciprocate generously. By transferring the full amount, I maximize the possible benefits from any future cooperation, embracing risk in the hope of triggering a shift that leads to high rewards. <10>"\\
        \makecell{o1-mini \\ (first round)} & As a risk-seeking assistant, I aim to maximize potential returns by transferring a larger portion of my endowment. By sending the full amount, the tripling effect amplifies the potential gain from the receiver's decision to return a portion, maximizing the overall payoff if the receiver reciprocates generously. <10>\\
        \makecell{o1-mini \\ (first round)}& As a risk-seeking player, I aim to maximize my potential gains by transferring a larger portion of my endowment. By sending more money, I increase the total amount the receiver has to potentially return, which could lead to a higher overall payoff for me despite the increased risk that the receiver might return less or nothing. Therefore, I choose to transfer the maximum amount available. <10>"\\
        \makecell{o1-mini \\ (later round)} & As a risk-seeking assistant, my strategy is to maximize potential returns despite previous outcomes. Even though the receiver returned \$0 in the last round, there is always a possibility that the receiver might respond differently in future interactions. By consistently transferring the maximum amount, I leverage the tripling mechanism to potentially gain the highest possible return if the receiver chooses to reciprocate. <10.00>" \\
        \makecell{DeepSeek-R1 \\(first round)} &  <think> Okay, so I need to figure out how much of my \$10 endowment to transfer to the receiver in each of the 10 rounds. Let me start by understanding the game mechanics again. First, I have \$10, and whatever amount I send gets tripled. So if I send \$X, the receiver gets \$3X. Then the receiver can choose to send back some amount Y from that \$3X. My final payment is (10 - X) + Y, and theirs is 10 + (3X - Y). Since this is repeated over 10 rounds, there's a possibility for building trust. If I send a higher amount initially, maybe the receiver will reciprocate by sending back more, leading to higher overall earnings for both of us in subsequent rounds. But since I'm risk-seeking, I might be more inclined to take chances to maximize potential returns. In the first round, without any history, the receiver has no incentive to send back money unless they're trustworthy or also playing a long-term strategy. But since we're playing 10 rounds, maybe they'll realize that being cooperative can lead to better mutual outcomes. If I send \$10, the receiver gets \$30. If they split it fairly, maybe they send back \$15, which would give me \$15 (since I kept \$0 and get \$15 back). But there's a risk they might keep all \$30, leaving me with nothing. However, if I'm risk-seeking, taking that chance could lead to higher rewards. Alternatively, sending a moderate amount like \$5 would result in \$15 for them. If they return half, that's \$7.50, so my total would be \$5 + \$7.50 = \$12.50. But again, they might not return anything. But over multiple rounds, if I start by sending a high amount and the receiver reciprocates, we can establish a pattern. If they don't reciprocate in the first round, I might reduce the amount in subsequent rounds. However, since I'm risk-seeking, maybe I should go all-in early to maximize potential gains, hoping the receiver will cooperate. Another angle: the receiver knows that if they don't return a decent amount, I might stop sending money in future rounds. So their rational choice might be to return a portion to keep me sending more. But how much would they return? Maybe a percentage like 1/3 of the tripled amount, which would be equal to the original sent amount. For example, if I send \$10, tripled to \$30, they return \$10, keeping \$20. Then my final payment is \$0 + \$10 = \$10, same as if I sent nothing. But if they return more, say \$15, I gain. But if I send \$10 and they return \$15, I get \$15, which is better than keeping the \$10. But they might not. So maybe the optimal strategy is to send a high amount initially, test the waters, and adjust based on their responses. Since it's 10 rounds, building trust could be beneficial. However, as a risk-seeker, I might prioritize the chance of higher returns over safety. So sending the full \$10 each round, hoping that the receiver will reciprocate enough to make it worthwhile. Even if they don't in some rounds, the potential upside in others could balance it out. Alternatively, maybe start with a high amount and adjust based on their returns. But since the user wants the answer for each round, and the question is about how much to send in each round considering the history, but since this is the first time, there's no history yet. So for the first round, the decision is based purely on strategy without prior interactions. Given that, a risk-seeking approach would be to send the maximum possible to maximize potential returns, hoping the receiver will reciprocate. Even if there's a chance of loss, the high risk could lead to high reward. So, I think the answer here is to send \$10 initially, then adjust in future rounds based on the receiver's responses. But since the question is about the first transfer, and considering the 10 rounds, starting with a high amount to establish cooperation. </think> As a risk-seeking sender, my strategy is to maximize potential returns by initially transferring the full amount to build trust and incentivize reciprocal generosity, adjusting in later rounds based on the receiver's responses.  <\$10> \\
        \makecell{DeepSeek-R1 \\(later round)} & <think> Okay, so I need to figure out how much to send to the receiver in this next round. Let me start by understanding the setup again. We both start with \$10 each. As the sender, I can send any amount from \$0 to \$10. Whatever I send gets tripled, so if I send \$10, the receiver gets \$30. Then the receiver decides how much to send back. But in the first round, I sent \$10, and they sent back \$0. That's pretty harsh. Now, this is the second round out of nine total, and the same person is the receiver each time. The problem is that if I keep sending \$10 and they keep returning \$0, I end up with nothing each time. My final payment is what I keep plus what they return. So if I send \$10, I keep \$0, and if they return \$0, I get \$0. Meanwhile, the receiver gets their \$10 plus the tripled amount minus what they return. In the first round, they ended up with \$10 + \$30 - \$0 = \$40. But since there are future rounds, maybe I can influence their behavior. If I stop sending them money, they might realize that they need to return some to encourage me to send more. But in the first round, I sent \$10, and they didn't return anything. So maybe they're not trustworthy. Wait, but this is a repeated game. In game theory, repeated interactions can lead to cooperation because players can punish each other for non-cooperative behavior. If I keep sending \$10 and they keep returning nothing, I might want to reduce the amount I send to punish them. But if I send less, maybe they'll start sending back more to incentivize me to send more again. But how many rounds are left? The user said there are 9 rounds total, and this is the second one. So after this, there are 7 more rounds. If I start sending \$0 now, the receiver might not have a chance to change their behavior. But maybe if I send a lower amount, like \$5, and see if they respond by sending something back. If they do, maybe I can increase again. If not, I might have to keep reducing. But the receiver's average return is \$0 from the first round. If they didn't send anything back when I sent \$10, maybe they're just selfish. But maybe they thought it was a one-time game. Now that it's repeated, they might realize that they need to return some to keep me sending money. Alternatively, maybe they're not thinking strategically and just taking the maximum for themselves each round. If that's the case, no matter what I send, they'll keep everything. So if I send \$10, they get \$30, keep it all, and I get nothing. If I send \$0, they can't take anything, so I keep my \$10. But if I send \$0, the receiver gets their \$10 plus \$0, so they get \$10. But if I send \$10, they get \$40. So from their perspective, they might prefer me sending \$10. But if they never return anything, I have no incentive to send anything. So maybe the optimal strategy is to send \$0 from now on. Because sending anything else just results in me losing money. But wait, maybe in the next rounds, if I send \$0, the receiver might start sending something to encourage me to send again. But since there are 7 more rounds after this, maybe there's a chance. Alternatively, maybe I should test them by sending a small amount. Like \$1. Then they get \$3, and maybe they'll return some part of it. If they return even \$1, then I get \$1 back, which is better than nothing. But if they return \$0 again, then I lost \$1. But given that in the first round they returned \$0 even when I sent \$10, they might not be cooperative. So maybe they're a defector. In that case, the best response is to send \$0 to minimize my losses. But let's think about possible strategies. If I send \$0 now, the receiver gets \$10 + \$0 = \$10. If I keep sending \$0, they can't get more than their \$10. But if I send some amount, maybe they'll start cooperating. But given their past behavior, it's risky. Alternatively, maybe I should try to send a smaller amount, like \$5. Then they get \$15 (tripled), and maybe they'll return some. If they return, say, \$5, then I get \$5 (from keeping \$5) + \$5 returned = \$10. Which is the same as if I kept all \$10. But if they return more, I could profit. \\
        \makecell{DeepSeek-R1 \\(continued)} &  But if they return less, I lose. But since they didn't return anything the first time, maybe they won't return anything again. So sending anything is a loss. Therefore, the safest bet is to send \$0. That way, I keep my \$10, and they can't take advantage of me. Even though it might not lead to cooperation in future rounds, at least I don't lose anything. But wait, maybe the receiver is using a tit-for-tat strategy. If I send \$10 and they send \$0, maybe they're just greedy. But if I send \$0 now, they might realize that they need to return something to get more in the future. But since there are 7 more rounds after this, maybe they'll start cooperating. Alternatively, maybe they're not going to change. So the best I can do is to maximize my own payoff by sending \$0. So, considering all this, I think sending \$0 is the best choice to avoid further losses. The receiver hasn't shown any willingness to cooperate, so trusting them again would be risky. Therefore, I should send \$0. </think> To minimize losses given the receiver's history of returning \$0 despite full transfers, and to avoid rewarding non-cooperative behavior, the optimal strategy is to send nothing. This ensures retaining the full \$10 endowment, as future reciprocity appears unlikely based on past actions.  <\$0.00> \\

    \caption{Reasoning Excerpts Generated from Different Models}
    \label{tab:model_reasoning}
\end{longtable}

\end{document}